\documentclass{article}

\PassOptionsToPackage{numbers}{natbib}
\usepackage[preprint]{neurips_2026}

\usepackage[utf8]{inputenc} 
\usepackage[T1]{fontenc}    
\usepackage{hyperref}       
\usepackage{url}            
\usepackage{booktabs}       
\usepackage{amsfonts}       
\usepackage{nicefrac}       
\usepackage{microtype}      
\usepackage{xcolor}         
\usepackage{graphicx}       
\usepackage{capt-of}
\usepackage{float}
\usepackage{enumitem}

\usepackage{amsmath}
\usepackage[ruled,vlined,linesnumbered]{algorithm2e}
\usepackage{wrapfig}
\setcitestyle{numbers,square}
\usepackage{makecell}
\usepackage[table]{xcolor}

\title{BrickAnything: Geometry-Conditioned Buildable Brick Generation with Structure-Aware Tokenization}
\author{
Zhengyang Ni\textsuperscript{1,2,3,*} \quad
Feng Yan\textsuperscript{1,2,3,*} \quad
Yu Guo\textsuperscript{1,2,3} \quad
Fei Wang\textsuperscript{1,2,3,\textdagger}
\\[1.5mm]
\textsuperscript{1}Xi'an Jiaotong University
\\
\textsuperscript{2}State Key Laboratory of Human-Machine Hybrid Augmented Intelligence
\\
\textsuperscript{3}Institute of Artificial Intelligence and Robotics
}
\begin{document}
\maketitle 

\begingroup
\renewcommand{\thefootnote}{\fnsymbol{footnote}}
\footnotetext[1]{Equal contribution.}
\footnotetext[2]{Corresponding author.}
\endgroup
\begin{abstract}
Generating physically buildable brick structures from 3D shapes requires more than geometric reconstruction: the output must also satisfy discrete part constraints and structural stability. Existing brick generation methods either rely on heuristic optimization, which can break down when the target 3D shape does not admit a feasible structure under predefined constraints, or generate brick sequences without explicitly modeling the underlying 3D geometry and assembly relations. 
In this work, we present \textbf{BrickAnything}, a geometry-conditioned autoregressive framework for generating buildable brick structures from diverse 3D representations. BrickAnything uses point clouds as a unified geometric interface and predicts brick sequences that reconstruct the target shape under assembly constraints. To model structural dependencies among bricks, we introduce a structure-aware tree tokenization, which represents brick structures through local attachment relations. This formulation makes sequence generation more consistent with the physical construction process,  and reduces invalid intermediate states. We further introduce preference-based alignment post-training, validity-constrained decoding and adaptive rollback to improve buildability objectives such as stability and geometric fidelity. Extensive experiments demonstrate that BrickAnything produces geometrically faithful and physically realizable brick structures, and that the proposed tokenization effectively reduces rollback and regeneration compared with conventional ordering strategies. Our code is available at \url{https://github.com/xjtunzy/BrickAnything}.
\end{abstract}

\section{Introduction}
Discrete brick-based structures offer a physically grounded and modular representation for constructing 3D objects, bridging digital geometry and real-world assembly.  Unlike conventional 3D representations such as mesh\cite{zhou2026mesh,wu2025direct3d,chen2024meshanything,zhao2025deepmesh,tang2024edgerunner,liu2025mesh}, point cloud\cite{vahdat2022lion,huang2024pointinfinity,hui2025not,ren2024tiger,sella2025blended,zhou2024frepolad}, Gaussian Splatting\cite{kerbl20233d,fan2024lightgaussian,Charatan_2024_CVPR,Jiang_2025_AnySplat}, or implicit fields\cite{mildenhall2021nerf,impl1,imp2,imp3}, brick assemblies are governed by strict combinatorial and structural constraints. A generated brick structure must not only approximate the target geometry, but also satisfy valid part compatibility, inter-brick connectivity, and structural stability. This combination of geometric fidelity and physical buildability makes brick generation a fundamentally different and more challenging problem than conventional 3D content synthesis.

Existing methods for brick structure generation can be broadly categorized into generative approaches and heuristic or search-based methods. Recent generative models, such as BrickGPT\cite{pun2025generating}, LEGO\textsuperscript{\textregistered}-Maker\cite{ge2025lego} and BrickNet\cite{kulits2026bricknet}, formulate brick construction as a sequential prediction problem, enabling flexible and expressive generation of complex assemblies. However, these methods typically rely on high-level inputs such as text or images, which do not provide explicit 3D geometric constraints. As a result, the generated structures are primarily guided by semantic priors or 2D visual cues, making it difficult to ensure accurate spatial configurations and faithful reconstruction of target shapes, especially for geometrically complex objects.
On the other hand, heuristic or search-based methods, including Image2Lego\cite{lennon2021image2lego} and Legolization\cite{luo2015legolization}, explicitly incorporate geometric and structural constraints by formulating brick assembly as a combinatorial optimization problem. While these methods can produce physically feasible structures, they often rely on fixed target representations and heuristic or search-based optimization strategies. This limits their flexibility when the target geometry is not directly realizable in the discrete brick space, and leads to high computational complexity as the search space grows.
This raises a natural question: \textbf{Can we design a unified framework for brick generation that explicitly leverages 3D geometry, adapts to diverse input representations, and generates buildable structures in a more flexible and effective manner?}

To address this question, we propose \textbf{BrickAnything}, a unified framework for geometry-conditioned buildable brick generation. We adopt point clouds as a modality-agnostic intermediate representation, allowing diverse 3D inputs to be mapped into a common geometric interface. Point clouds provide explicit spatial information while being easily derived from a wide range of 3D modalities, making them a natural interface for geometry-conditioned generation.

Conditioned on this explicit 3D geometry, we formulate brick construction as an autoregressive sequence generation problem. A key component of BrickAnything is a structure-aware tree tokenization scheme. Instead of representing brick assemblies only as spatially ordered sequences, our tokenization organizes bricks according to local attachment relations. This representation better reflects the physical construction process, produces more compact sequences, and makes structural dependencies between bricks explicit during generation.

Building on this tokenization, we further propose a buildability-aware reward that jointly measures structural stability and geometric fidelity. 
This reward is used to construct preference pairs for Direct Preference Optimization (DPO), enabling the model to optimize buildability-related objectives that are difficult to capture with likelihood-based training alone. During inference, we also introduce a lightweight parent-aware rollback strategy tailored to the proposed tree tokenization. When an unstable brick is detected, the generation process rolls back to its parent brick in the tree sequence and regenerates the subsequent structure. This simple mechanism effectively improves the robustness of autoregressive brick generation.

Through extensive experiments, we demonstrate that BrickAnything improves both geometric fidelity and buildability compared with conventional brick generation baselines. In particular, our structure-aware tokenization explicitly preserves local attachment relations and substantially reduces rollback and regeneration during inference. We summarize our contributions as follows:

\begin{itemize}[leftmargin=*, itemsep=0.1em]
    \item We introduce \textbf{BrickAnything}, a unified geometry-conditioned framework for buildable brick generation. By using point clouds as a modality-agnostic intermediate representation, BrickAnything enables explicit 3D-guided generation from diverse input modalities.

    \item We propose a structure-aware tree tokenization scheme for brick structures. Instead of relying only on global spatial ordering, our representation organizes bricks according to local attachment relations, explicitly modeling parent-child assembly dependencies during autoregressive generation.

    \item We propose a buildability-aware reward that jointly measures structural stability and geometric fidelity, and use it to construct preference pairs for Direct Preference Optimization (DPO), enabling the model to optimize buildability-related objectives beyond likelihood-based training.

    \item We design a lightweight parent-aware rollback strategy tailored to the proposed tree tokenization. When an unstable brick is detected, the generation process rolls back to its structurally related parent and regenerates the subsequent substructure, improving inference robustness.
\end{itemize}

\section{Related work}

\subsection{Brick Structure Generation}
Generating brick structures from a reference 3D shape has been widely studied in prior work\cite{luo2015legolization,testuz2013automatic}.
Early approaches formulate this task as a combinatorial construction problem guided by handcrafted heuristics or search strategies. These methods typically enforce constraints such as structural connectivity, stability, and efficient brick usage, and often rely on sequential planning or heuristic search to assemble bricks that reconstruct a target geometry. While effective in producing physically feasible designs, they generally assume that the input shape can be directly realized using discrete bricks. As a result, they fail when there is no feasible assembly solution for the given brick set — i.e., when the target geometry cannot be constructed under the prescribed brick types, connection rules, and stability constraints. Moreover, the reliance on heuristic rules or search-based strategies leads to high computational costs and limited scalability.

Recent learning-based methods have explored generative brick modeling with autoregressive models\cite{pun2025generating,kulits2026bricknet,ge2025lego}.
BrickGPT\cite{pun2025generating} generates physically stable and buildable brick structures from text prompts by predicting bricks sequentially, while enforcing validity and physical constraints during inference.
BrickNet~\citep{kulits2026bricknet} further extends autoregressive brick generation by introducing a graph-backed connectivity representation and a large-scale human-designed LDraw dataset, enabling the modeling of more diverse part types and connection semantics.
LEGO\textsuperscript{\textregistered}-Maker\cite{ge2025lego} further explores image-conditioned LEGO\textsuperscript{\textregistered} model generation using an autoregressive framework with multiple brick types, but its experiments mainly focus on a limited set of object categories.
Despite these advances, existing learning-based methods are mainly conditioned on text descriptions or 2D visual observations, which do not provide explicit and unambiguous 3D geometric constraints.
As a result, they may struggle to faithfully reconstruct complex target geometries or generate structurally valid brick structures that are tightly aligned with an input 3D shape.
This motivates a unified geometry-conditioned formulation that directly uses 3D shape information while preserving the buildability constraints of brick assemblies.

\subsection{Autoregressive Models for 3D Generation}
Autoregressive models have become an effective paradigm for modeling structured data by factorizing generation into sequential token prediction\cite{vaswani2017attention}. Following their success in language\cite{radford2019language,brown2020language} and image generation\cite{parmar2018image,chen2020generative,tian2024visual}, this paradigm has also been explored for 3D generation across different representations, including point cloud\cite{cheng2022autoregressive,meng20253d}, Gaussian splatting\cite{von2026gaussiangpt,liu2026avatarpointillist}, and mesh\cite{nash2020polygen,siddiqui2024meshgpt}. 
Brick-based generation can also be naturally formulated as a sequential prediction problem, since a brick assembly of brick structures consists of discrete elements with spatial, topological, and physical dependencies. Recent methods\cite{pun2025generating,ge2025lego,kulits2026bricknet} serialize brick assemblies into token sequences and generate brick structures step by step. In this work, we introduce a geometry-conditioned autoregressive framework that uses point clouds as explicit 3D guidance and organizes brick assemblies according to local attachment relations, enabling more accurate and physically buildable generation.

\subsection{RLHF with Direct Preference Optimization}
Aligning model outputs with human preferences is a critical step in improving generative models, especially for large language models (LLMs)\citep{llm1,llm2,llm3}. Techniques like reinforcement learning from human feedback (RLHF) and direct preference optimization (DPO) have become standard for incorporating preference signals. While early RLHF methods using PPO\citep{llm4} often suffer from instability and high computational cost, DPO\cite{llm5dpo} eliminates the need for an explicit reward model and directly optimizes policy likelihood. Recently, preference‑based optimization has also been applied to structured generation tasks, including 3D shape generation\citep{zhao2025deepmesh,zhou2025dreamdpo,guo2025autoconnect}. In this work, we adapt DPO to brick structure generation by constructing preference pairs with a buildability-aware reward that jointly considers geometric fidelity and structural stability.

\section{Method}
\begin{figure}[t]
    \centering
    \includegraphics[width=\linewidth]{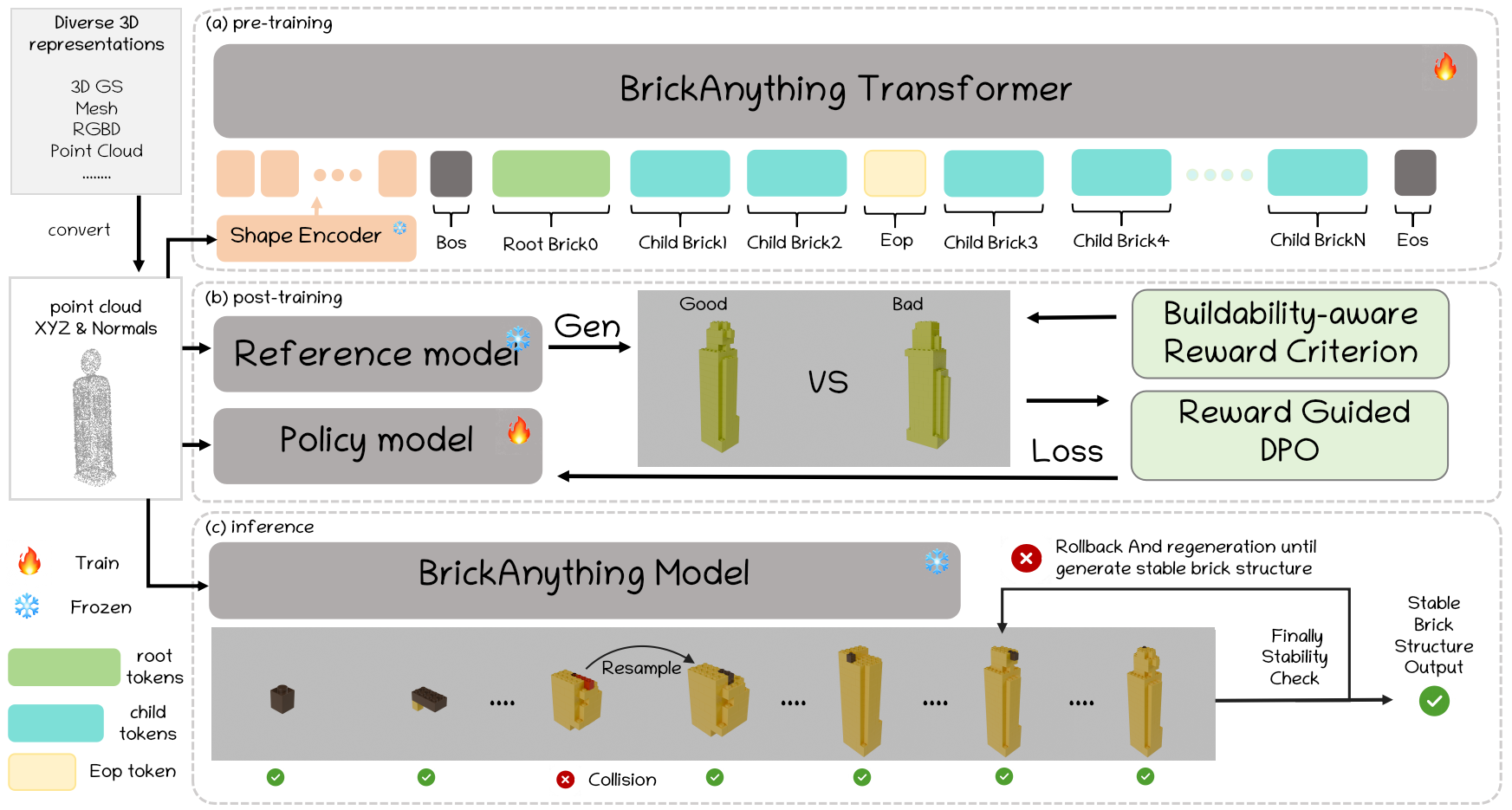}
    \caption{\textbf{BrickAnything Framework Overview.} 
    BrickAnything converts diverse 3D inputs into point clouds as a modality-agnostic geometric interface. 
    The framework consists of three stages: 
    1) \textbf{Pre-training}, where a shape encoder and BrickAnything Transformer learn structure-aware autoregressive brick generation; 
    2) \textbf{Post-training}, where buildability-aware rewards are used to construct preference pairs and optimize the policy model with reward-guided DPO; and 
    3) \textbf{Inference}, where validity-constrained decoding and rollback-based regeneration are applied to produce stable brick structures.}
    \label{fig:pipeline}
\end{figure}

BrickAnything comprises three components, as illustrated in Figure~\ref{fig:pipeline}.
Following BrickGPT~\citep{pun2025generating}, we construct brick structures within a \(20\times20\times20\) voxel grid using the same eight commonly available standard LEGO\textsuperscript{\textregistered} bricks: 
\(1 \times 1\), \(1 \times 2\), \(1 \times 4\), \(1 \times 6\), \(1 \times 8\), \(2 \times 2\), \(2 \times 4\), and \(2 \times 6\).
Section~\ref{sec:tokenization} introduces our tokenization scheme and model architecture.
Section~\ref{sec:dpo} presents a buildability-aware DPO post-training stage, where preference pairs are constructed by jointly considering geometric fidelity and structural stability.
Section~\ref{sec:roll} introduces validity-constrained decoding and stability-guided rollback.

\subsection{Structure-Aware Tokenization and Geometry-Conditioned Generation}
\label{sec:tokenization}
Unlike prior brick generation methods~\citep{pun2025generating,ge2025lego} that represent structures as flattened sequences that preserve spatial order, we propose a structure-aware tree tokenization scheme that explicitly encodes local attachment relations between bricks.


\begin{figure}[t]
    \centering
    \includegraphics[width=\linewidth]{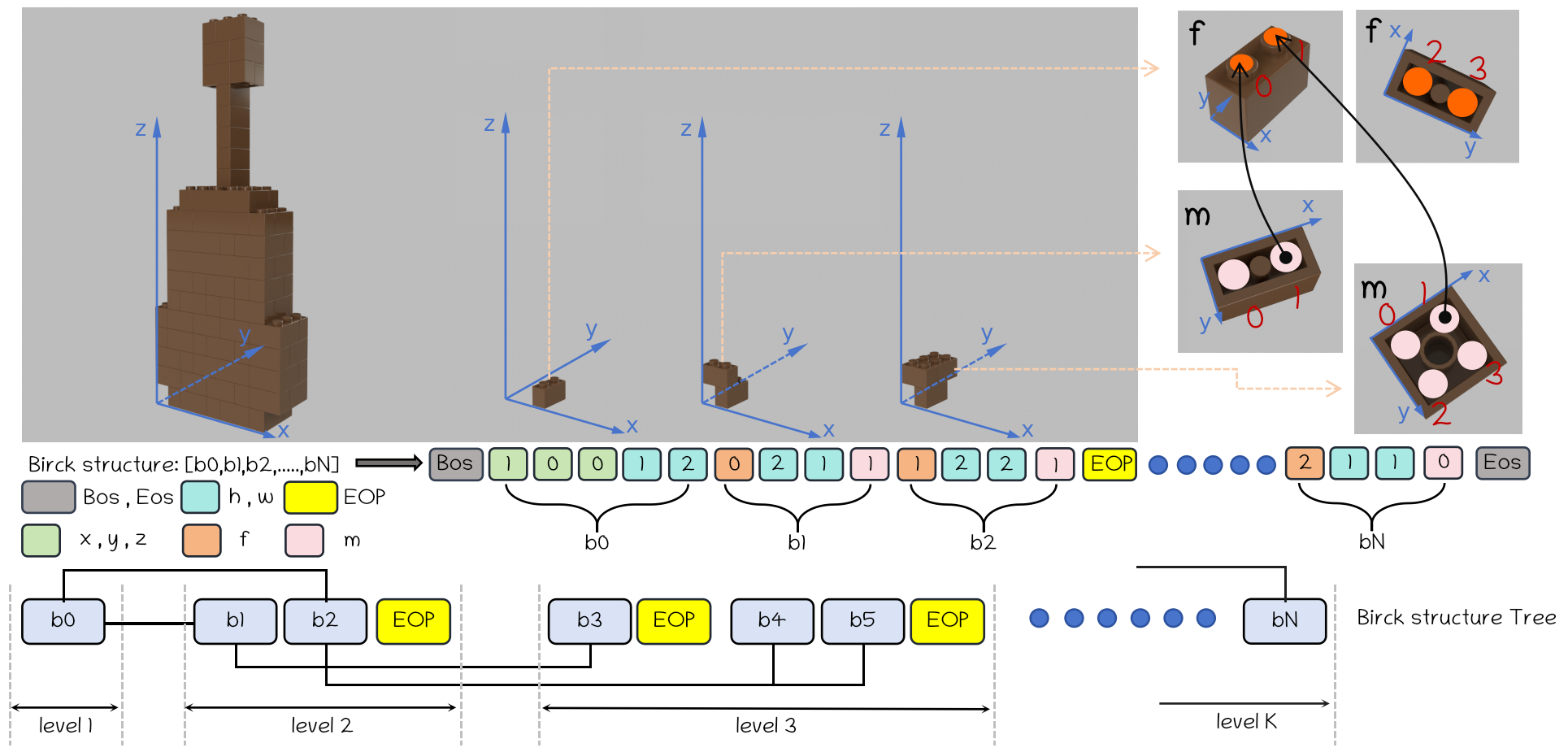}
    \caption{Overview of the proposed structure-aware tree tokenization. 
The root brick is encoded by absolute attributes \((x,y,z,h,w)\), while each child brick is encoded by relative attachment tokens \((f,h,w,m)\). 
\texttt{EOP} marks the end of each parent’s child group in the BFS traversal.}
    \label{fig:tokenization}
\end{figure}

\paragraph{Structure-aware tree tokenization.}
Given a brick structure \(\mathcal{B}=\{b_i\}_{i=1}^{N}\), each brick is represented as
\(b_i=[h_i,w_i,x_i,y_i,z_i]\), where \((h_i,w_i)\) denotes its footprint size along the
\(x\)- and \(y\)-axes, and \((x_i,y_i,z_i)\) is the discrete position of the stud closest
to the origin. We first construct a vertical attachment graph
\(\mathcal{G}=(\mathcal{B},\mathcal{E})\), where an edge indicates that two bricks are
vertically adjacent and overlap in the \(xy\)-plane:
\begin{equation}
    (b_i,b_j)\in\mathcal{E}
    \quad \Longleftrightarrow \quad
    |z_i-z_j|=1
    \ \text{and}\
    \Omega_i^{xy}\cap \Omega_j^{xy}\neq \emptyset 
\end{equation}
where \(\Omega_i^{xy}\) is the footprint of \(b_i\) on the \(xy\)-plane. 
For a stable brick structure, \(\mathcal{G}\) is connected~\cite{luo2015legolization}.
We therefore choose the root brick by the lexicographic order of \((z,y,x)\) and perform
breadth-first traversal over \(\mathcal{G}\). Each unvisited brick is assigned to the
first visited neighboring parent, yielding a deterministic BFS spanning tree, as shown in
Figure~\ref{fig:tokenization}.

The root brick is encoded by its absolute attributes:
\begin{equation}
    (x_0,y_0,z_0,h_0,w_0)
\end{equation}
For each non-root brick \(b_i\), we encode it relative to its parent \(b_{p(i)}\) as
\begin{equation}
    (f_i,h_i,w_i,m_i)
\end{equation}
where \(f_i\) indexes the attachment position on the parent, \((h_i,w_i)\) is the child
brick size, and \(m_i\) indexes the child-side anchor position, as shown in Figure~\ref{fig:tokenization}.
During BFS traversal, children of the same parent are sorted increasingly by \(f_i\).
After all children of the current parent are encoded, an \texttt{EOP} token is inserted
to mark the end of this local attachment group. The final sequence is
\begin{equation}
\mathbf{s}=
[\texttt{BOS},
x_0,y_0,z_0,h_0,w_0,
(f,h,w,m),\ldots,\texttt{EOP},
(f,h,w,m),\ldots,\texttt{EOP},
\ldots,
\texttt{EOS}]
\end{equation}
This representation explicitly attaches each newly generated brick to an existing parent,
thereby converting brick generation from global coordinate prediction into local
attachment prediction. More details of the proposed tokenization scheme are provided in Appendix~\ref{more-details-of-tokenization}.

\paragraph{Shape-conditioned brick generation.}
Given an input 3D shape, we first convert it into a point cloud representation with surface normals, denoted as 
\(\mathcal{P}\in\mathbb{R}^{N\times 3}\) and \(\mathcal{N}\in\mathbb{R}^{N\times 3}\), where \(N=8192\).
The point cloud and its normals are encoded by a pre-trained Michelangelo\citep{zhao2023michelangelo} encoder \(\mathcal{E}_{g}\), producing a sequence of shape tokens
\begin{equation}
\mathbf{c}_{g}=\mathcal{E}_{g}(\mathcal{P},\mathcal{N})
\end{equation}
which serves as the geometric condition for brick structure generation. 
We choose OPT-350M\citep{zhang2022opt} as our autoregressive transformer architecture. 
Given the structure-aware tree sequence 
$\mathbf{s}=(s_1,s_2,\ldots,s_T)$, 
our transformer decoder with parameters \(\theta\) generates the sequence conditioned on the shape tokens \(\mathbf{c}_g\). 
The model is trained using the standard next-token prediction objective:
\begin{equation}
\mathcal{L}_{\mathrm{SFT}}
=
-\sum_{i=1}^{T}
\log p_{\theta}
\left(
s_i \mid s_{<i}, \mathbf{c}_g
\right)
\end{equation}
During inference, generation starts from the shape condition and the \texttt{BOS} token, and 
the generated sequence is then converted into a brick structure through our tree detokenization algorithm. 

\subsection{DPO-based Post-Training for Buildable Brick Generation}
\label{sec:dpo}
Next-token prediction optimizes token-level likelihood but cannot directly enforce non-differentiable objectives such as geometric fidelity and physical buildability. 
To address this, we introduce a reward-guided DPO post-training stage. 
Specifically, for each input shape, we generate multiple candidate brick structures and rank them according to our proposed buildability-aware reward. 
The ranked candidates are then used to fine-tune the model with a reward-weighted DPO objective, together with an auxiliary SFT loss.

\paragraph{Buildability-aware reward.}
\label{iouandcd}
Given an input point cloud \(\mathcal{P}\) and a generated brick structure 
\(\hat{\mathcal{B}}\), we evaluate generation quality from two aspects: 
geometric fidelity and physical stability. 
For geometric fidelity, we define a reward that combines voxel-level occupancy 
consistency and surface-level geometric alignment.

We first voxelize the input point cloud into an occupancy grid 
\(V_{\mathcal{P}}\), and convert the generated brick structure into another 
occupancy grid \(V_{\hat{\mathcal{B}}}\). 
The voxel-level consistency is measured by the intersection-over-union:
\begin{equation}
  R_{\mathrm{IoU}}
=
\frac{|V_{\mathcal{P}}\cap V_{\hat{\mathcal{B}}}|}
{|V_{\mathcal{P}}\cup V_{\hat{\mathcal{B}}}|}
\in [0,1]  
\end{equation}
Although IoU measures volumetric overlap, it may overlook fine-grained surface discrepancies.
Therefore, we introduce a surface-level distance metric to complement the voxel-level score and better capture boundary alignment and local geometric details.

Specifically, we extract a surface mesh \(\hat{\mathcal{M}}\) from the generated
voxel grid \(V_{\hat{\mathcal{B}}}\) using Marching Cubes~\citep{we1987marching}, and
uniformly sample a point cloud \(\hat{\mathcal{P}}\) from the reconstructed surface.
Before computing the distance, both \(\hat{\mathcal{P}}\) and \(\mathcal{P}\) are centered at their centroids and scaled by their maximum radial distances.
We then compute the Chamfer Distance between the normalized point clouds:
\begin{equation}
D_{\mathrm{CD}}(\mathcal{P}, \hat{\mathcal{P}})
=
\frac{1}{|\mathcal{P}|}
\sum_{\mathbf{x}\in \mathcal{P}}
\min_{\hat{\mathbf{x}}\in \hat{\mathcal{P}}}
\|\mathbf{x}-\hat{\mathbf{x}}\|_2
+
\frac{1}{|\hat{\mathcal{P}}|}
\sum_{\hat{\mathbf{x}}\in \hat{\mathcal{P}}}
\min_{\mathbf{x}\in \mathcal{P}}
\|\hat{\mathbf{x}}-\mathbf{x}\|_2 
\end{equation}
A smaller Chamfer Distance indicates better surface-level alignment between the 
generated brick structure and the target geometry. 
We convert this distance into a bounded reward by
\begin{equation}
R_{\mathrm{CD}}
=
\max\left(1-5D_{\mathrm{CD}}(\mathcal{P},\hat{\mathcal{P}}), 0\right)
\in [0,1]
\end{equation}
The final geometric reward is defined as the sum of the voxel-level and 
surface-level rewards:
\begin{equation}
R_{\mathrm{geo}}
=
R_{\mathrm{IoU}} + R_{\mathrm{CD}}
\in [0,2]
\end{equation}
For structural stability, we follow StableLego\citep{liu2024stablelego} and estimate a per-brick stability score \(s_i\in[0,1]\) based on static equilibrium under frictional constraints. 
Since a single unstable brick can compromise the whole assembly, we define the stability reward as the minimum per-brick score:
\begin{equation}
R_{\mathrm{stable}}
=
\min_{b_i\in\hat{\mathcal{B}}} s_i
\in [0,1]
\end{equation}
The overall reward is therefore
\begin{equation}
R(\hat{\mathcal{B}},\mathcal{P})
=
R_{\mathrm{geo}} + R_{\mathrm{stable}}
\in [0,3]
\end{equation}

\begin{figure}[t]
    \centering
    \includegraphics[width=\linewidth]{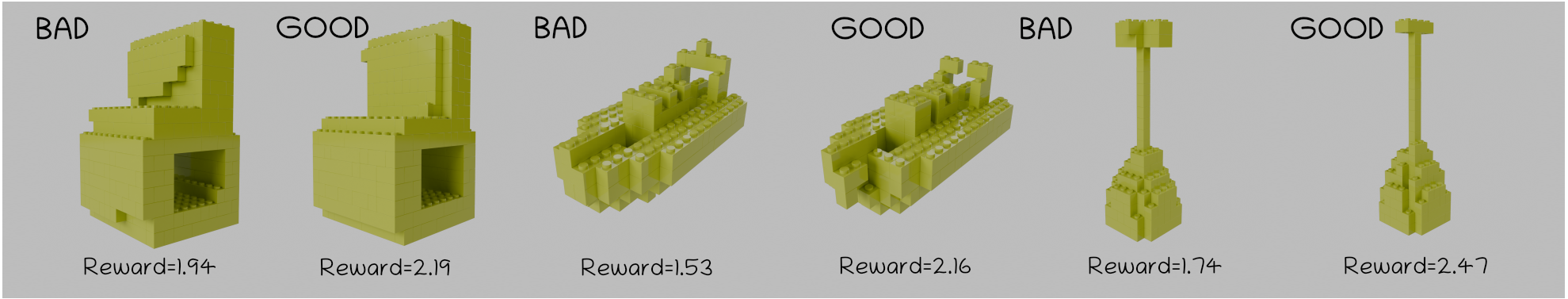}
    \caption{Examples of the collected preference pairs for reward-guided DPO.}
    \label{fig:dposet}
\end{figure}

\paragraph{Preference pair construction.}
For each input shape, we sample six legal candidate brick structures from the SFT model and compute their rewards. 
We construct pairwise comparisons and retain only informative pairs whose reward gap is at least \(0.2\) and whose higher reward is no less than \(1\). 
For each retained pair, the higher-reward candidate is used as the preferred sample \(y_w\), and the other as the rejected sample \(y_l\).
Figure~\ref{fig:dposet} shows some selection cases of our collected preference pairs.

\paragraph{Reward-guided DPO.}
Given an input condition \(x\), a preferred sequence \(y_w\), and a rejected sequence \(y_l\), we optimize the policy model \(\pi_\theta\) with a reward-weighted DPO objective:
\begin{equation}
\mathcal{L}_{\mathrm{DPO}}
=
-\mathbb{E}_{(x,y_w,y_l)}
\left[
\Delta R
\log \sigma
\left(
\beta
\log
\frac{\pi_\theta(y_w|x)\pi_{\mathrm{ref}}(y_l|x)}
{\pi_{\mathrm{ref}}(y_w|x)\pi_\theta(y_l|x)}
\right)
\right]
\end{equation}
where \(\pi_{\mathrm{ref}}\) is the frozen reference model, \(\beta\) controls the strength of the KL constraint, and
\begin{equation}
\Delta R = R(y_w,x)-R(y_l,x)
\end{equation}
The reward gap \(\Delta R\) assigns larger weights to pairs with clearer quality differences, encouraging the model to distinguish high-fidelity and stable assemblies from inferior ones.

To preserve the original data distribution, we further add an auxiliary SFT loss on the ground-truth sequence \(y_{\mathrm{gt}}\):
\begin{equation}
\mathcal{L}_{\mathrm{SFT}}
=
-\sum_{t=1}^{T}
\log \pi_\theta(y_{\mathrm{gt},t}\mid y_{\mathrm{gt},<t},x)
\end{equation}
The final post-training objective is
\begin{equation}
\mathcal{L}_{\mathrm{post}}
=
\mathcal{L}_{\mathrm{DPO}}
+
\lambda \mathcal{L}_{\mathrm{SFT}}
\end{equation}
where \(\lambda\) balances preference alignment and distribution preservation.

\begin{figure}[t]
    \centering
    \includegraphics[width=\linewidth]{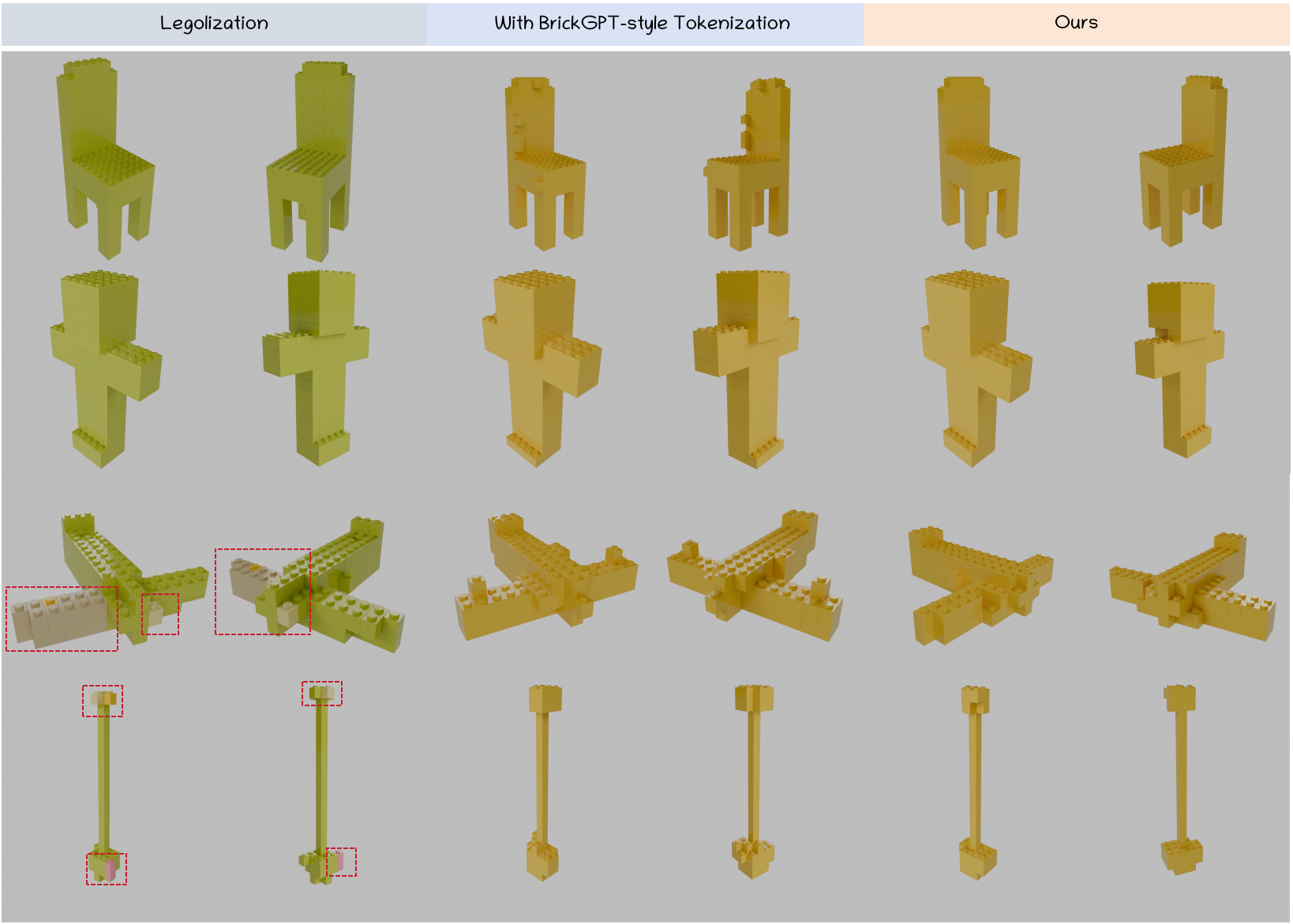}
    \caption{\textbf{Qualitative comparison.}
BrickAnything produces stable and geometry-faithful structures than the baselines.
Red boxes mark Legolization failures.}
    \label{fig:qualitative_comparison}
\end{figure}
\subsection{Validity-Constrained Decoding and Stability-Guided Rollback}
\label{sec:roll}
\paragraph{Validity-constrained decoding.}
Autoregressive decoding may produce invalid local decisions, including unsupported brick sizes, invalid attachment tokens, and spatial collisions.
Instead of accepting each decoded child tuple directly, we perform tuple-level validity checking during inference.
For each child brick, the model samples a candidate tuple \((f,h,w,m)\), which is decoded into a concrete brick placement based on the current parent brick and BFS traversal state.
The candidate tuple is then verified in two stages.
First, we check whether the token combination defines a valid attachment, including a feasible parent connector \(f\), a brick size \((h,w)\) from the predefined brick library, and a valid child-side anchor \(m\).
Second, after the candidate is converted into grid occupancy, we check whether it overlaps with any previously generated brick.
If either check fails, the candidate tuple is rejected and resampled.
The accepted tuple is appended to the sequence only after passing these validity checks.
\paragraph{stability-guided rollback.}
Although validity-constrained decoding filters invalid local decisions, it does not guarantee the global physical stability of the complete structure.
To further improve physical buildability, we perform stability-guided rollback after a complete brick structure is generated.
Following StableLego~\citep{liu2024stablelego}, we compute a per-brick stability score \(s_i\in[0,1]\) for each brick in the generated structure, where \(s_i=0\) denotes that brick \(b_i\) is unstable.
If the generated structure contains unstable bricks, we identify the first unstable brick according to the generation order:
\begin{equation}
   k=\min\{i\mid s_i=0\} 
\end{equation}
We then trace its parent brick $b_{p(k)}$ in the generated tree structure and locate the token position where $b_{p(k)}$ itself is generated in the generated token sequence.
The sequence is rolled back to the state before generating $b_{p(k)}$ and the subsequent tokens are regenerated.
This process is repeated until a stable structure is obtained or the maximum rollback budget is reached.
The parent-child attachment relations allow our strategy to locate the source of instability more precisely, regenerate the affected substructure, and preserve stable preceding bricks, leading to more targeted and structurally consistent correction.

\definecolor{best}{RGB}{255,190,190}
\definecolor{second}{RGB}{255,220,180}
\definecolor{third}{RGB}{255,255,180}

\begin{table}[t]
\centering
\caption{\textbf{Quantitative comparison with baselines and ablations.}
We compare BrickAnything with representative baselines and ablated variants on the challenging and stable subsets.
Lower CD and rollback averages are better, while higher IoU, \%Stable, and \%Valid indicate better performance.
\textcolor{best}{\rule{1.2em}{0.8em}},
\textcolor{second}{\rule{1.2em}{0.8em}}, and
\textcolor{third}{\rule{1.2em}{0.8em}}
indicate the best, second-best, and third-best results, respectively.}
\label{tab:quantitative_analysis}
\setlength{\tabcolsep}{6pt}
\renewcommand{\arraystretch}{1.12}

\textbf{(a) Challenging subset}
\vspace{0.3em}

\resizebox{\linewidth}{!}{
\begin{tabular}{lccccc}
\toprule
\textbf{Method}
& \textbf{CD $\downarrow$}
& \textbf{IoU $\uparrow$}
& \makecell{\textbf{Rollback $\downarrow$}}
& \textbf{\%Stable $\uparrow$}
& \textbf{\%Valid $\uparrow$} \\
\midrule
Legolization\citep{luo2015legolization}
& -- 
& -- 
& -- 
& 0.0\% 
& \cellcolor{best}100\% \\
\midrule
BrickGPT\citep{pun2025generating}-style tokenization w/o DPO
& 0.1307 
& 0.544 
& 6.991 
& 74.0\% 
& \cellcolor{best}100\% \\

BrickGPT\citep{pun2025generating}-style tokenization
& 0.1309 
& 0.553 
& \cellcolor{third}6.750 
& \cellcolor{third}76.0\% 
& \cellcolor{best}100\% \\

\midrule

Ours w/o Validity-constrained decoding or Rollback or DPO
& \cellcolor{best}0.1271
& \cellcolor{best}0.602
& -- 
& 60.6\% 
& 68.6\% \\

Ours w/o Rollback or DPO
& \cellcolor{second}0.1283 
& \cellcolor{second}0.597 
& -- 
& 72.0\% 
& \cellcolor{best}100\% \\

Ours w/o DPO
& 0.1301 
& 0.573 
& \cellcolor{second}0.444 
& \cellcolor{second}82.4\% 
& \cellcolor{best}100\% \\

\textbf{Ours}
& \cellcolor{third}0.1299 
& \cellcolor{third}0.586 
& \cellcolor{best}0.422 
& \cellcolor{best}83.4\%
& \cellcolor{best}100\% \\
\bottomrule
\end{tabular}
}

\vspace{0.9em}

\textbf{(b) Stable subset}
\vspace{0.3em}

\resizebox{\linewidth}{!}{
\begin{tabular}{lccccc}
\toprule
\textbf{Method}
& \textbf{CD $\downarrow$}
& \textbf{IoU $\uparrow$}
& \makecell{\textbf{Rollback $\downarrow$}}
& \textbf{\%Stable $\uparrow$}
& \textbf{\%Valid $\uparrow$} \\
\midrule
Legolization\citep{luo2015legolization}
& \cellcolor{best}0.1120
& \cellcolor{best}1.000
& -- 
& \cellcolor{best}100\% 
& \cellcolor{best}100\% \\

\midrule
BrickGPT\citep{pun2025generating}-style tokenization w/o DPO
& 0.1297 
& 0.721 
& 3.916 
& \cellcolor{best}100\% 
& \cellcolor{best}100\% \\

BrickGPT\citep{pun2025generating}-style tokenization
& 0.1292 
& 0.742 
& 3.620 
& \cellcolor{best}100\% 
& \cellcolor{best}100\% \\
\midrule

Ours w/o Validity-constrained decoding or Rollback or DPO
& \cellcolor{second}0.1265 
& \cellcolor{second}0.798 
& -- 
& 63.4\% 
& 71.6\% \\

Ours w/o Rollback or DPO
& \cellcolor{third}0.1274 
& 0.772 
& -- 
& 91.0\% 
& \cellcolor{best}100\% \\

Ours w/o DPO
& 0.1290 
& 0.751 
& \cellcolor{second}0.216 
& \cellcolor{best}100\% 
& \cellcolor{best}100\% \\

\textbf{Ours}
& 0.1283 
& \cellcolor{third}0.788 
& \cellcolor{best}0.184 
& \cellcolor{best}100\%
& \cellcolor{best}100\% \\
\bottomrule
\end{tabular}
}

\vspace{-0.5em}
\end{table}
\section{Experiments}
\subsection{Implementation Details}
\paragraph{Dataset.}
We construct our training data from a curated collection of approximately 230K high-quality meshes from ShapeNet\citep{chang2015shapenet}, Objaverse\citep{deitke2023objaverse}, and Objaverse-XL\citep{deitke2023objaverse-xl}, following recent mesh curation practice~\citep{chen2025meshanything}.
Applying Legolization~\citep{luo2015legolization} yields around 168K stable mesh--brick pairs, of which 4K are randomly selected for validation and the rest are used for training.
For evaluation, we build two test subsets: a stable subset with 500 validation samples successfully converted by Legolization, and a challenging subset with 500 meshes sampled from cases where Legolization fails to produce stable structures. 
More training details are provided in Appendix~\ref{sec:training}.

\subsection{Metrics and Baselines}
\paragraph{Metrics}
We evaluate generated brick structures in terms of geometric fidelity, structural validity, physical stability, and inference correction cost.
Geometric fidelity is measured by Voxel IoU and Chamfer Distance, as defined in Sec.~\ref{iouandcd}; structural validity is the percentage of outputs satisfying valid brick types, workspace bounds, and collision-free placement; physical stability is measured by the stable generation rate following StableLego~\citep{liu2024stablelego}; and inference cost is measured by the average number of rollback operations.
Unless otherwise specified, Voxel IoU, Chamfer Distance, and rollback average are computed over final generated samples that are physically stable.
\paragraph{Baselines.}
We compare our method with two representative baselines.
First, we implement a baseline using the same point-cloud encoder and transformer architecture as ours, but replacing the proposed structure-aware tree tokenization with a conventional BrickGPT~\citep{pun2025generating}-style tokenization.
Specifically, each brick is represented as \((h,w,x,y,z)\), and bricks are sorted into a flat sequence by the lexicographic order of \((z,y,x)\).
The baseline is pretrained on the same dataset as our model and equipped with the same brick-by-brick rejection sampling and physics-aware rollback strategies used in BrickGPT~\citep{pun2025generating}, isolating the effect of the proposed tokenization scheme.
Second, we compare with Legolization~\citep{luo2015legolization}, a heuristic search-based method for converting 3D shapes into stable brick structures.
This comparison evaluates whether our learned model can produce geometrically faithful and physically buildable structures, especially in cases where heuristic search fails.

\begin{figure}[t]
   \centering
    \includegraphics[width=\linewidth]{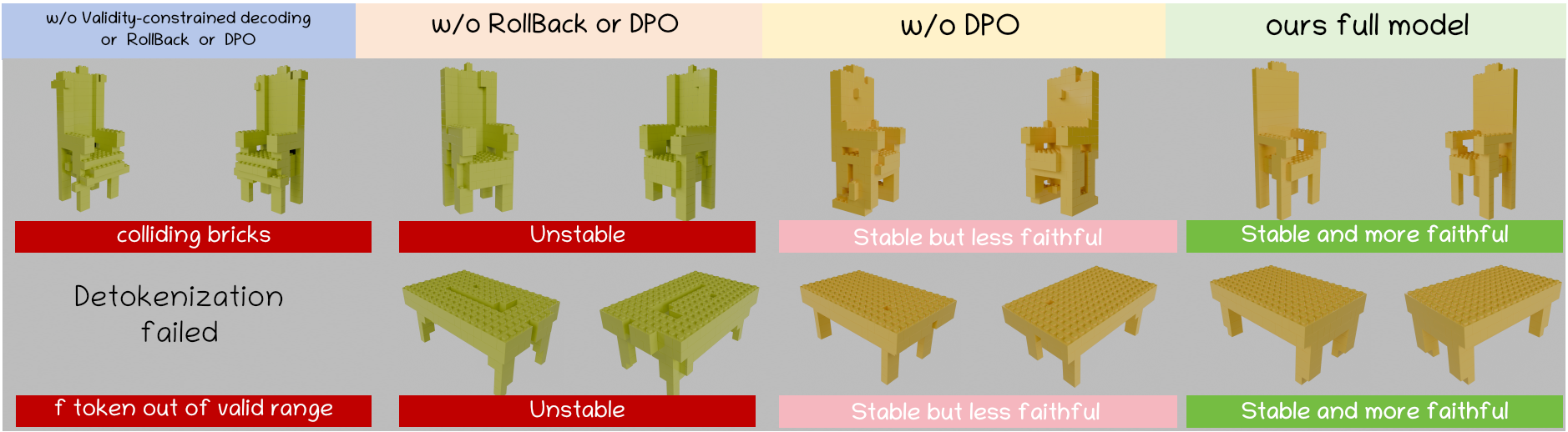}
    \caption{\textbf{Qualitative ablation results.}
Each component improves buildability or fidelity: validity-constrained decoding removes invalid placements, rollback corrects instability, and DPO enhances shape faithfulness.}
    \label{fig:abla}
\end{figure}

\subsection{Comparison}

Table~\ref{tab:quantitative_analysis} compares BrickAnything with representative baselines on the challenging and stable subsets.
On the challenging subset, Legolization~\citep{luo2015legolization} obtains \(0.0\%\) stable rates, indicating that heuristic search is brittle for geometrically difficult targets.
The BrickGPT-style baseline~\citep{pun2025generating} benefits from DPO, improving the stable rate from \(74.0\%\) to \(76.0\%\), IoU from \(0.544\) to \(0.553\), and reducing the average number of rollbacks from \(6.991\) to \(6.750\).
In contrast, BrickAnything achieves stronger buildability and fidelity, reaching \(83.4\%\) stable rate, \(100\%\) valid rate, \(0.586\) IoU, and a much lower rollback average of \(0.422\).

On the stable subset, Legolization obtains the best CD and IoU because this subset is selected from its successful cases.
Among learned methods, BrickAnything achieves the best overall performance.
Compared with the BrickGPT-style baseline, BrickAnything improves IoU from \(0.742\) to \(0.788\), reduces CD from \(0.1292\) to \(0.1283\), and lowers the rollback average from \(3.620\) to \(0.184\), while maintaining \(100\%\) stable and valid rates.
Together with Figure~\ref{fig:qualitative_comparison}, these results show that structure-aware tree tokenization improves geometric fidelity and sequence coherence, while DPO, validity-constrained decoding, and rollback enhance buildability with fewer correction steps.
Additional qualitative results, failure analyses, and image- and text-to-brick extensions are provided in Appendix~\ref{sec:mmmore_results}.

\subsection{Ablation Studies}

Table~\ref{tab:quantitative_analysis} and Figure~\ref{fig:abla} ablate the main components of BrickAnything.
Removing validity-constrained decoding, rollback, and DPO leads to strong geometric scores but substantially weaker buildability, with the stable and valid rates dropping to \(60.6\%\) and \(68.6\%\) on the challenging subset.
Adding validity-constrained decoding restores the valid rate to \(100\%\), but the stable rate remains limited without rollback, indicating that local validity constraints alone cannot guarantee global physical stability.
Rollback further improves stability by correcting unstable generations through structurally related resampling.
Finally, DPO improves the generation distribution itself: compared with the variant without DPO, the full model improves IoU from \(0.573\) to \(0.586\) on the challenging subset and from \(0.751\) to \(0.788\) on the stable subset, while reducing the average number of rollbacks from \(0.444\) to \(0.422\) and from \(0.216\) to \(0.184\), respectively.
These results show that validity-constrained decoding, rollback, and DPO address complementary aspects of buildable brick generation.
We provide a detailed ablation analysis in Appendix~\ref{sec:detailed_ablation}.
\section{Conclusion}
We presented \textbf{BrickAnything}, a geometry-conditioned framework for buildable brick generation.
By combining point-cloud conditioning with structure-aware tree tokenization, BrickAnything models local attachment relations in autoregressive generation.
Buildability-aware DPO, validity-constrained decoding, and stability-guided rollback further improve fidelity and stability.
Experiments demonstrate superior performance over coordinate-ordered tokenization baselines and robustness when heuristic Legolization fails.
We discuss the limitations of our work and potential directions for future research in Appendix~\ref{sec:limitations}.

\medskip

{
\small
\bibliographystyle{unsrtnat} 
\bibliography{references}         
}

\newpage
\appendix
\section*{Appendix}

\section{More Details of Structure-Aware Tree Tokenization}
\label{more-details-of-tokenization}
\subsection{Attachment Token Definition and Detokenization}
\begin{figure}[H]
   \centering
    \includegraphics[width=\linewidth]{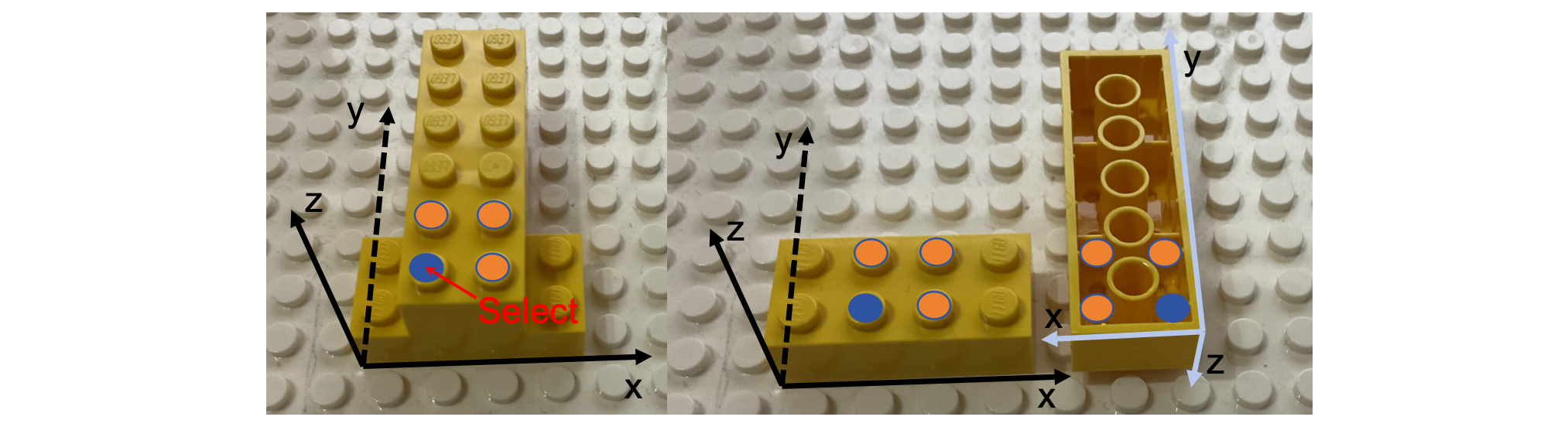}
    \caption{\textbf{Example of attachment token definition.}
    The selected shared stud is expressed in the local coordinate systems of both the parent and child bricks.
    The parent-side local coordinate determines \(f_i\), while the child-side local coordinate determines \(m_i\).}
    \label{fig:att}
\end{figure}
For each non-root brick \(b_i\), we encode its attachment to the parent brick \(b_{p(i)}\) using a deterministic reference stud within their overlapping footprint.
Let \(\Omega_{p(i)}^{xy}\) and \(\Omega_i^{xy}\) denote the sets of occupied stud locations of the parent and child bricks in the global \(xy\)-plane.
Since a valid vertical attachment requires
\(\Omega_{p(i)}^{xy}\cap\Omega_i^{xy}\neq\emptyset\),
we select a canonical attachment location
\(
(q_i^x,q_i^y)
=
\min_{}
\left(
\Omega_{p(i)}^{xy}\cap\Omega_i^{xy}
\right)
\)
where \(\min\) denotes the lexicographically smallest stud location.
This deterministic choice removes ambiguity when multiple studs overlap.
We then express this global attachment location in the local coordinate systems of the parent and child bricks:
\begin{equation}
    (u_i^p,v_i^p) = (q_i^x-x_{p(i)},\, q_i^y-y_{p(i)})
\end{equation}
\vspace{-3em}
\begin{equation}
    (u_i^c,v_i^c) = (q_i^x-x_i,\, q_i^y-y_i)
\end{equation}

The parent-side attachment token \(f_i\) encodes the vertical direction and the local attachment position on the parent brick, while the child-side anchor token \(m_i\) encodes the corresponding local anchor position on the child brick:
\begin{equation}
    f_i = s_i h_{p(i)}w_{p(i)} + v_i^p h_{p(i)} + u_i^p
\end{equation}
\vspace{-3em}
\begin{equation}
    m_i = v_i^c h_i + u_i^c
\end{equation}
where \(s_i\in\{0,1\}\) indicates the vertical attachment direction, with \(s_i=0\) meaning that the child is attached above the parent and \(s_i=1\) meaning that it is attached below the parent.

Figure~\ref{fig:att} shows a simple example.
The selected shared stud has local coordinate \((u_i^p,v_i^p)=(1,0)\) on the parent brick and \((u_i^c,v_i^c)=(0,0)\) on the child brick.
Since the child is attached above the parent, we have \(s_i=0\).
For a parent brick with \((h_{p(i)},w_{p(i)})=(4,2)\) and a child brick with \(h_i=2\), the corresponding tokens are \(f_i=0\times2\times4+0\times4+1=1\) and \(m_i=0\times2+0=0\).

During detokenization, given the parent state and the child size \((h_i,w_i)\), we first recover the local attachment variables from \(f_i\) and \(m_i\): 
\(s_i=\left\lfloor f_i/(h_{p(i)}w_{p(i)})\right\rfloor\), 
\(r_i=f_i\bmod (h_{p(i)}w_{p(i)})\), 
\(u_i^p=r_i\bmod h_{p(i)}\), 
\(v_i^p=\left\lfloor r_i/h_{p(i)}\right\rfloor\), 
\(u_i^c=m_i\bmod h_i\), and 
\(v_i^c=\left\lfloor m_i/h_i\right\rfloor\).
The child position is then recovered as 
\(x_i=x_{p(i)}+u_i^p-u_i^c\), 
\(y_i=y_{p(i)}+v_i^p-v_i^c\), and 
\(z_i=z_{p(i)}+(1-2s_i)\).
Thus, the attachment tokens provide a reversible local representation of each child brick relative to its parent.
\subsection{Structure-aware tree tokenization and detokenization algorithm}
\vspace{-0.5em}
\begin{algorithm}[H]
\caption{Structure-aware Tree Tokenization}
\label{alg:tree_tokenization}
\KwIn{A brick assembly $\mathcal{B}=\{b_i\}_{i=1}^{N}$, where $b_i=(x_i,y_i,z_i,h_i,w_i)$}
\KwOut{A token sequence $\mathbf{s}$}

Construct a vertical attachment graph $\mathcal{G}=(\mathcal{B},\mathcal{E})$\;
\ForEach{pair of bricks $(b_i,b_j)$}{
    \If{$|z_i-z_j|=1$ and $\Omega_i^{xy}\cap\Omega_j^{xy}\neq\emptyset$}{
        Add edge $(b_i,b_j)$ to $\mathcal{E}$\;
    }
}

Select root brick $b_{root}=\arg\min_{b_i}(z_i,y_i,x_i)$\;
Initialize queue $Q\leftarrow[b_{root}]$\;
Initialize visited set $\mathcal{V}\leftarrow\{b_{root}\}$\;
Initialize token sequence
$\mathbf{s}\leftarrow[\texttt{BOS},x_0,y_0,z_0,h_0,w_0]$\;

\While{$Q$ is not empty}{
    Pop current parent brick $b_p$ from $Q$\;
    Find unvisited neighbors $\mathcal{C}(b_p)$ of $b_p$ in $\mathcal{G}$\;
    Sort $\mathcal{C}(b_p)$ by the parent-side connector index $f$\;
    
    \ForEach{child brick $b_c\in\mathcal{C}(b_p)$}{
        Compute parent-side connector index $f$\;
        Compute child-side anchor index $m$\;
        Append $(f,h_c,w_c,m)$ to $\mathbf{s}$\;
        Add $b_c$ to $\mathcal{V}$\;
        Push $b_c$ into $Q$\;
    }
    
    Append $\texttt{EOP}$ to $\mathbf{s}$\;
}

Remove redundant trailing $\texttt{EOP}$ tokens\;
Append $\texttt{EOS}$ to $\mathbf{s}$\;
\Return $\mathbf{s}$\;
\end{algorithm}
\vspace{-0.5em}
\begin{algorithm}[H]
\caption{Structure-aware Tree Detokenization}
\label{alg:tree_detokenization}
\KwIn{A token sequence $\mathbf{s}$}
\KwOut{A reconstructed brick assembly $\hat{\mathcal{B}}$}

Construct root brick $\hat{b}_{root}=(x_0,y_0,z_0,h_0,w_0)$\;
Initialize $\hat{\mathcal{B}}\leftarrow\{\hat{b}_{root}\}$\ and queue $Q\leftarrow[\hat{b}_{root}]$\;

\While{$\mathbf{s}$ still has remaining tokens}{
    \If{$Q$ is empty}{
        \textbf{break}\;
    }

    Pop current parent brick $\hat{b}_p$ from $Q$\;
    Initialize child list $\mathcal{C}\leftarrow\emptyset$\;

    \While{next token is not $\texttt{EOP}$ and tokens remain}{
        Read child tuple $(f,h_c,w_c,m)$\;
        Decode parent-side connector $(s_f,u_p,v_p)$ from $f$\;
        Decode child-side anchor $(u_c,v_c)$ from $m$\;

        Recover child horizontal position:
        $x_c\leftarrow x_p+u_p-u_c$,
        $y_c\leftarrow y_p+v_p-v_c$\;

        \If{$s_f=0$}{
            Set $z_c\leftarrow z_p+1$\;
        }
        \Else{
            Set $z_c\leftarrow z_p-1$\;
        }

        Construct child brick $\hat{b}_c=(x_c,y_c,z_c,h_c,w_c)$\;
        Add $\hat{b}_c$ to $\hat{\mathcal{B}}$ and $\mathcal{C}$\;
    }

    \If{next token is $\texttt{EOP}$}{
        Consume $\texttt{EOP}$\;
    }

    Push all bricks in $\mathcal{C}$ into $Q$\;
}

\Return $\hat{\mathcal{B}}$\;
\end{algorithm}
\subsection{Structure-Aware Tree Tokenization Codebook Size and Length Analysis}

\paragraph{Codebook size.}
We compare the codebook size of the conventional BrickGPT\citep{pun2025generating}-style tokenization and our structure-aware tree tokenization.
For the BrickGPT-style baseline, each brick is represented by five tokens, e.g. \((h,w,x,y,z)\).
The coordinate tokens share a discrete resolution of \(20\), corresponding to coordinate values in the \(20\times20\times20\) workspace.
The brick-size tokens share five possible values, \(\{1,2,4,6,8\}\), and three additional special tokens are used for \texttt{BOS}, \texttt{EOS}, and \texttt{PAD}.
Thus, the BrickGPT-style codebook size is
\begin{equation}
    |\mathcal{V}_{\mathrm{BrickGPT}}|
=
20+5+3
=
28
\end{equation}

Our structure-aware tree tokenization uses the same coordinate tokens, brick-size tokens, and special tokens.
In addition, it introduces one \texttt{EOP} token, parent-side attachment tokens \(f\), and child-side anchor tokens \(m\).
The \(f\)-token set is determined by the maximum parent footprint area and two vertical attachment directions, while the \(m\)-token set is determined by the maximum child footprint area.
Given our brick library
\begin{equation}
    \{1\times1,1\times2,1\times4,1\times6,1\times8,2\times2,2\times4,2\times6\}
\end{equation}

the maximum footprint area is
\begin{equation}
A_{\max}=2\times6=12
\end{equation}
Therefore, the \(f\)-token set has size \(2A_{\max}=24\), and the \(m\)-token set has size \(A_{\max}=12\).
The total codebook size of our method is
\begin{equation}
|\mathcal{V}_{\mathrm{ours}}|
=
20+5+3+1+24+12
=
65
\end{equation}

\paragraph{Sequence length.}
Let \(N\) be the number of bricks in a structure.
In the BrickGPT-style tokenization, each brick is represented by five tokens.
Ignoring padding, the sequence length is
\begin{equation}
T_{\mathrm{BrickGPT}} = 5N+2
\end{equation}
where the two extra tokens correspond to \texttt{BOS} and \texttt{EOS}.

In our structure-aware tree tokenization, the root brick is encoded by five absolute tokens \((x_0,y_0,z_0,h_0,w_0)\), while each non-root brick is encoded by four relative tokens \((f,h,w,m)\).
In principle, an \texttt{EOP} token can be inserted after each parent's child group.
However, in our experiments, we omit redundant \texttt{EOP} tokens for leaf nodes with no children, since there is no child group to terminate.
Let \(I\) denote the number of non-leaf parent nodes that emit \texttt{EOP}.
The sequence length of our tokenization is then
\begin{equation}
T_{\mathrm{ours}}
=
1+5+4(N-1)+I+1
=
4N+I+3
\end{equation}

where the first and last terms correspond to \texttt{BOS} and \texttt{EOS}.
Since \(I\leq N-1\), we have
\begin{equation}
T_{\mathrm{ours}}\leq 5N+2
\end{equation}

which shows that our representation is no longer than the coordinate-ordered representation, while explicitly encoding parent-child attachment relations.

\begin{table}[t]
\centering
\caption{\textbf{Codebook size and sequence length comparison.}
The reported average sequence length is computed on the training set.}
\label{tab:tokenization_efficiency_appendix}
\setlength{\tabcolsep}{8pt}
\renewcommand{\arraystretch}{1.1}
\begin{tabular}{lcc}
\toprule
Tokenization & Codebook size & Avg. sequence length \(\downarrow\) \\
\midrule
BrickGPT-style tokenization & 28 & 1050 \\
Structure-aware tree tokenization & 65 & \textbf{1044} \\
\bottomrule
\end{tabular}
\end{table}

As shown in Table~\ref{tab:tokenization_efficiency_appendix}, our tokenization uses a larger codebook due to the additional attachment tokens, but achieves a slightly shorter average sequence length.

\section{Additional Results and Analysis}
\label{sec:mmmore_results}

\subsection{Additional Qualitative Results}

\begin{figure}[H]
    \centering
    \includegraphics[width=\linewidth]{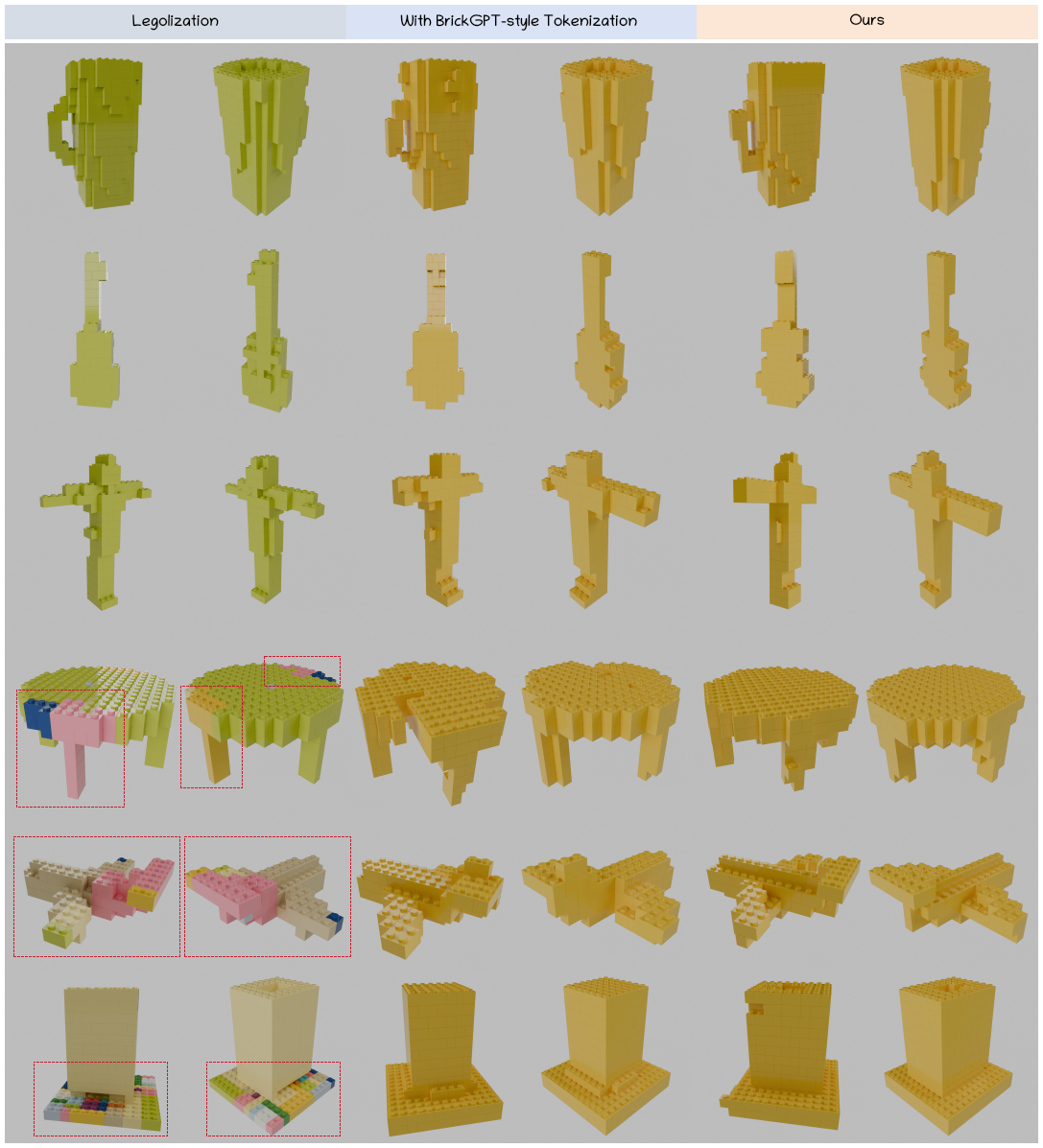}
    \caption{\textbf{Additional qualitative results.}
    We provide more examples of BrickAnything on diverse input shapes.
    The generated brick structures preserve the overall geometry of the input point clouds while satisfying brick-level construction constraints.}
    \label{fig:more_results}
\end{figure}

Figure~\ref{fig:more_results} shows additional generation results on diverse input shapes.
BrickAnything generally preserves the global shape structure while producing physically buildable brick assemblies.
\vspace{-2em}

\subsection{Failure Cases}
\begin{figure}[H]
    \centering
    \includegraphics[width=\linewidth]{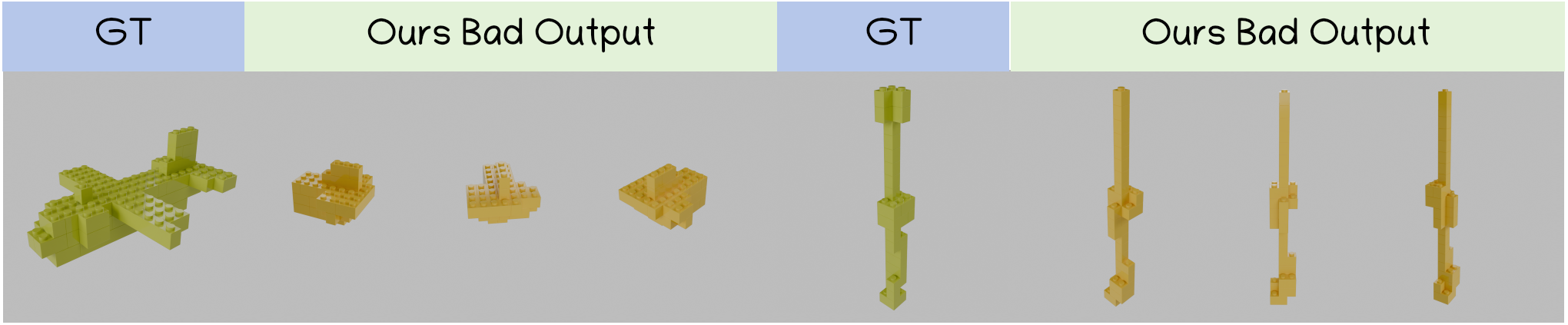}
    \caption{\textbf{Failure cases.}
    We show representative failure cases of BrickAnything.
    }
    \label{fig:failure_cases}
\end{figure}

Figure~\ref{fig:failure_cases} shows representative failure cases.
Failures mainly occur for inputs whose geometry depends on slender components, smooth curved surfaces, or subtle local details, where the limited workspace resolution and restricted brick library make accurate reconstruction difficult.

\subsection{Image- and Text-to-Brick Results}

\begin{figure}[H]
    \centering
    \includegraphics[width=\linewidth]{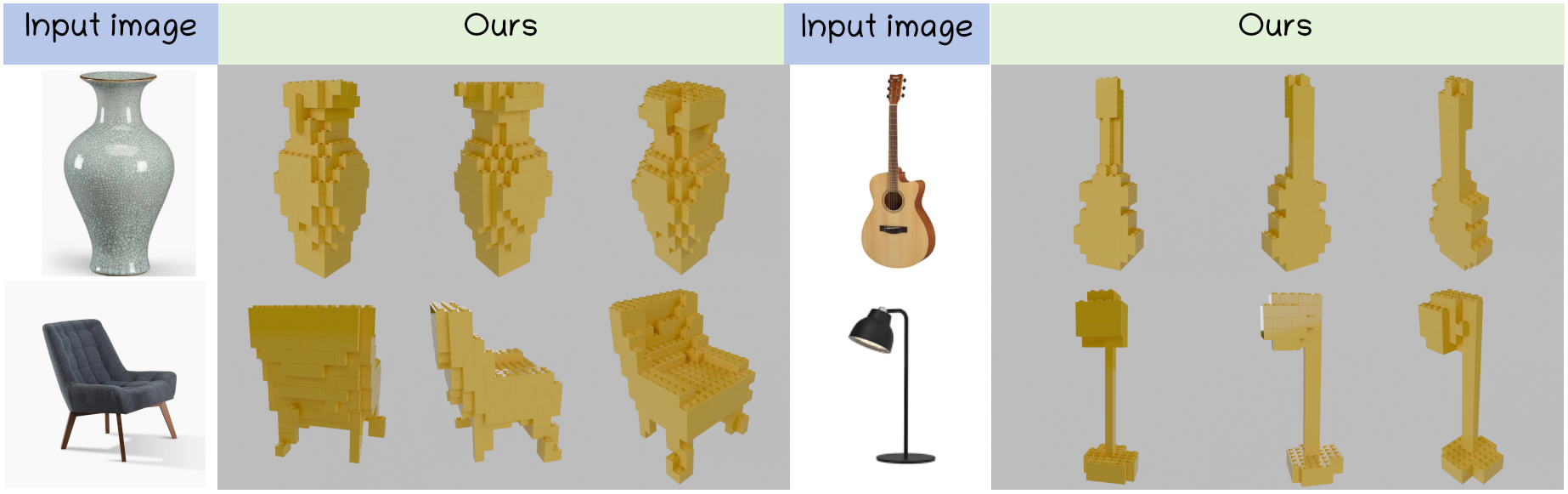}
    \caption{\textbf{Image-to-brick results.}
    Given an input image, we first use Hunyuan3D 2.5\citep{lai2025hunyuan3d} to reconstruct a 3D mesh, and then convert the mesh into a point cloud with normals as the input condition for BrickAnything.
    The generated brick structures preserve the major geometry of the input objects while satisfying brick-level construction constraints.}
    \label{fig:img2brick_results}
\end{figure}

\begin{figure}[H]
    \centering
    \includegraphics[width=\linewidth]{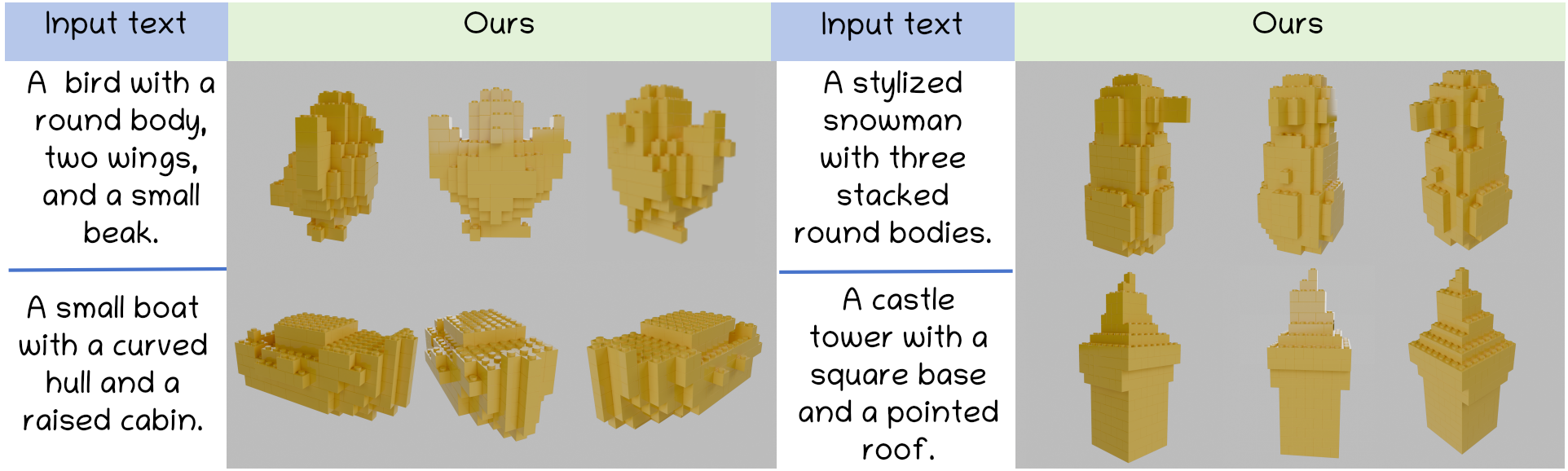}
    \caption{\textbf{Text-to-brick results.}
    Given an input text prompt, we first use Hunyuan3D 2.5\citep{lai2025hunyuan3d} to generate a corresponding 3D mesh, and then convert it into a point cloud with normals as the geometric condition for BrickAnything.
    The results show that BrickAnything can be extended to text-conditioned brick generation through an intermediate 3D representation.}
    \label{fig:text2brick_results}
\end{figure}
Although BrickAnything takes point clouds as input, it can be naturally extended to image- and text-conditioned brick generation by using an external 3D generation model as a geometry converter.
Specifically, we feed the input image or text prompt into Hunyuan3D 2.5\citep{lai2025hunyuan3d} to obtain a corresponding 3D mesh.
The mesh is then converted into a point cloud with normals, which serves as the geometric condition for BrickAnything.
This pipeline allows BrickAnything to generate buildable brick structures from higher-level modalities while still relying on explicit 3D geometric guidance.

\section{Training Details}
\label{sec:training}

We pre-train BrickAnything on 8 NVIDIA A100 80GB GPUs with a batch size of 16 per GPU, giving a global batch size of 128.
The model is trained for 50 epochs, taking approximately one day.
For each mesh, we randomly sample 8192 surface points with normal vectors as the input shape representation.
The autoregressive decoder is trained with fp16 mixed precision using AdamW.
We use a base learning rate of \(1\times10^{-4}\), a final learning rate of \(6\times10^{-5}\), weight decay \(0.1\), cosine scheduling, first-epoch warmup from \(1\times10^{-6}\), and gradient clipping with maximum norm 1.0.
We also apply random scaling, shifting, and rotation augmentations during training.

For DPO post-training, we sample 4K training shapes and construct 16K curated preference pairs using the proposed buildability-aware reward.
The pre-sampled point clouds corresponding to each mesh are directly loaded during DPO training, without additional mesh resampling.
We fine-tune the model for 10 epochs on 8 NVIDIA A100 80GB GPUs with a per-GPU batch size of 8.
The DPO coefficient is set to \(\beta=1\), and the auxiliary SFT loss weight is set to \(\lambda=1\).
We use a base learning rate of \(5\times10^{-7}\), a final learning rate of \(1\times10^{-7}\), a warmup learning rate of \(1\times10^{-7}\), gradient clipping of 10.0, and reserve 10\% of the preference pairs for validation.

\section{Detailed Ablation Analysis}
\label{sec:detailed_ablation}

We provide a detailed analysis of the ablated variants in Table~\ref{tab:quantitative_analysis} and Figure~\ref{fig:abla}.

\paragraph{Effect of validity-constrained decoding.}
Validity-constrained decoding is critical for ensuring that the generated token sequence can be converted into a legal brick structure.
When validity-constrained decoding, rollback, and DPO are all removed, the model achieves strong geometric scores, reaching \(0.1271\) CD and \(0.602\) IoU on the challenging subset.
However, its buildability is substantially weakened, with the stable and valid rates dropping to \(60.6\%\) and \(68.6\%\), respectively.
A similar trend is observed on the stable subset, where this variant obtains \(0.1265\) CD and \(0.798\) IoU, but only reaches \(63.4\%\) stable rate and \(71.6\%\) valid rate.
This indicates that high geometric overlap alone does not guarantee legal or physically realizable brick assemblies.
As shown in Figure~\ref{fig:abla}, unconstrained generation may produce invalid token combinations or spatial collisions, which can further lead to detokenization failures.

\paragraph{Effect of rollback.}
Adding validity-constrained decoding restores the valid rate to \(100\%\) on both subsets, confirming its effectiveness in preventing illegal placements.
However, without rollback, the stable rate remains limited: \(72.0\%\) on the challenging subset and \(91.0\%\) on the stable subset.
This shows that local validity constraints can ensure legal brick placements, but they cannot fully guarantee global physical stability.
Rollback complements validity-constrained decoding by identifying unstable generated structures and resampling structurally related brick placements.
With rollback, the variant without DPO improves the stable rate from \(72.0\%\) to \(82.4\%\) on the challenging subset, and from \(91.0\%\) to \(100\%\) on the stable subset.
Thus, validity-constrained decoding mainly addresses structural legality, while rollback is necessary for correcting stability failures.

\paragraph{Effect of buildability-aware DPO.}
DPO improves both the BrickGPT-style baseline and our structure-aware tree-tokenized model.
For the BrickGPT-style baseline, DPO improves the stable rate from \(74.0\%\) to \(76.0\%\) on the challenging subset, increases IoU from \(0.544\) to \(0.553\), and reduces the average number of rollbacks from \(6.991\) to \(6.750\).
On the stable subset, where both variants already achieve \(100\%\) stable and valid rates, DPO still improves IoU from \(0.721\) to \(0.742\) and reduces rollback average from \(3.916\) to \(3.620\).

DPO also benefits our tree-tokenized model.
Compared with the variant without DPO, the full model improves IoU from \(0.573\) to \(0.586\) on the challenging subset and from \(0.751\) to \(0.788\) on the stable subset.
It also reduces the average number of rollbacks from \(0.444\) to \(0.422\) and from \(0.216\) to \(0.184\), respectively.
These results suggest that buildability-aware preference optimization improves the generation distribution itself, leading to more geometry-faithful structures and reducing the need for rollback-based correction.

\paragraph{Effect of structure-aware tree tokenization.}
Compared with the BrickGPT-style coordinate-ordered tokenization, our structure-aware tree tokenization consistently improves fidelity, stability, and correction efficiency.
Without DPO, our model improves the challenging-subset IoU from \(0.544\) to \(0.573\), increases the stable rate from \(74.0\%\) to \(82.4\%\), and reduces rollback average from \(6.991\) to \(0.444\).
With DPO, BrickAnything further improves over the BrickGPT-style baseline, increasing IoU from \(0.553\) to \(0.586\), improving the stable rate from \(76.0\%\) to \(83.4\%\), and reducing rollback average from \(6.750\) to \(0.422\).
On the stable subset, BrickAnything also improves IoU from \(0.742\) to \(0.788\) and reduces rollback average from \(3.620\) to \(0.184\), while maintaining \(100\%\) stable and valid rates.
These results indicate that explicitly modeling local attachment relations yields more coherent brick sequences and substantially reduces the correction burden during inference.

\paragraph{Summary.}
Overall, the ablation results show that the three components address complementary aspects of buildable brick generation.
Validity-constrained decoding ensures local legality, rollback corrects global stability failures, and buildability-aware DPO improves the model distribution toward more faithful and stable generations.
In addition, structure-aware tree tokenization provides a stronger sequence representation than coordinate-ordered tokenization, leading to higher fidelity, higher stability, and fewer rollback corrections.
\section{Limitations and Future Work}
\label{sec:limitations}

BrickAnything is currently evaluated in a restricted \(20\times20\times20\) workspace with eight standard brick types.
While this setting enables controlled comparison, it also constrains the resolution, part diversity, and geometric complexity of the generated structures.
The failure cases in Fig.~\ref{fig:failure_cases} further suggest that our current representation is less effective for shapes whose geometry relies on slender components, subtle local variations, or smooth continuous surfaces.
In such cases, the coarse voxelized workspace and limited set of rectangular brick primitives may force the model to produce over-simplified or structurally distorted approximations.
Future work will extend the framework to larger workspaces, richer brick libraries, and more diverse connection types.

Our training data is generated by applying Legolization to curated mesh collections, and incorporating human-designed brick models could improve data diversity and construction realism.
In addition, our buildability-aware reward is hand-designed from geometric fidelity and stability metrics.
Learning more adaptive reward models or incorporating real assembly feedback may further improve physical buildability.



\end{document}